\pdfoutput=1

\documentclass[11pt]{article}

\usepackage{acl}

\usepackage{times}
\usepackage{latexsym}

\usepackage[T1]{fontenc}

\usepackage[utf8]{inputenc}

\usepackage{microtype}

\usepackage{amsmath,graphicx}
\usepackage{booktabs}
\usepackage{multirow}
\usepackage{xcolor}
\usepackage{amsfonts}
\usepackage[inline]{enumitem}
\usepackage{thmtools}
\usepackage{mathtools}
\usepackage{amsthm}
\usepackage{amssymb}
\usepackage{pgfplots}
\usepackage{tikz}
\usetikzlibrary{arrows,automata,positioning}
\usepackage{caption}
\usepackage{subcaption}
\usepackage{relsize}
\usepackage{url}
\usepackage{color, colortbl}
\usepackage{makecell}

\title{Text-Free Prosody-Aware Generative Spoken Language Modeling}

\author{
Eugene Kharitonov$^*$, Ann Lee$^*$, Adam Polyak, Yossi Adi, \\
{\bf Jade Copet, Kushal Lakhotia, Tu-Anh Nguyen, Morgane Rivière,} \\
{\bf Abdelrahman Mohamed, Emmanuel Dupoux, Wei-Ning Hsu} \\
Facebook AI Research\\
\texttt{\{kharitonov,annl,wnhsu\}@fb.com}
}

\begin{document}
\maketitle
\begin{abstract}
Speech pre-training has primarily demonstrated efficacy on classification tasks, while its capability of generating novel speech, similar to how GPT-2 can generate coherent paragraphs, has barely been explored. Generative Spoken Language Modeling (GSLM)~\cite{Lakhotia2021} is the only prior work addressing the generative aspects of speech pre-training, which replaces text with discovered phone-like units for language modeling and shows the ability to generate meaningful novel sentences. Unfortunately, despite eliminating the need of text, the units used in GSLM discard most of the prosodic information. Hence, GSLM fails to leverage prosody for better comprehension, and does not generate expressive speech. 
In this work, we present a prosody-aware generative spoken language model (pGSLM). It is composed of a multi-stream transformer language model (MS-TLM) of speech, represented as discovered unit and prosodic feature streams, and an adapted HiFi-GAN model converting MS-TLM outputs to waveforms. We devise a series of metrics for prosody modeling and generation, and re-use metrics from GSLM for content modeling. Experimental results show that the pGSLM can utilize prosody to improve both prosody and content modeling, and also generate natural, meaningful, and coherent speech given a spoken prompt.\footnote{Audio samples can be found at \url{https://speechbot.github.io/pgslm/}. Codes and models are available at \url{https://github.com/pytorch/fairseq/tree/main/examples/textless_nlp/pgslm}.}
\end{abstract}
\newcommand{\lf}{lf}
\newcommand{\todo}[1]{{\color{red}#1}}

\newcommand{\red}[1]{{\color{red}#1}}
\newcommand{\new}[1]{{\color{blue}#1}} 

\newcommand\eugene[1]{\textcolor{green!60!black}{EUGENE: #1}}
\newcommand\wn[1]{\textcolor{blue!80!black}{WN: #1}}
\newcommand\dpx[1]{\textcolor{orange!60!black}{dpx: #1}}

\section{Introduction}
%
Natural language processing (NLP) has made tremendous progress recently. One of the most significant findings is that language models (LMs) are natural unsupervised multitask learners~\cite{radford2018improving, radford2019language, brown2020language} --- by simply training a big neural network on next word prediction with a large amount of unlabeled text, it learns to comprehend, answer questions, summarize, and even translate~\cite{radford2019language}. Fine-tuning such pre-trained models further leads to the state-of-the-art performance on numerous benchmark tasks~\cite{brown2020language}, beating tailor-made models trained from scratch only on labeled data.

Given the impressive performance of pre-trained text language models, it is tempting to approach spoken language processing tasks by first transcribing speech into text with an automatic speech recognition (ASR) system and then utilizing text-based models for comprehension and generation. However, there are a number of caveats for such a framework. 
First, the majority of the world's languages are primarily spoken and do not have associated texts in large quantities~\cite{ethnologue}. In practice, this limits the reach of NLP techniques to a fraction of the world's languages that have a large presence on the web and for which there exists a widely available high quality ASR system.
Second, despite sharing the same vocabulary and syntactic rules, the spoken form and the written form of the same language still vary significantly in terms of sentence lengths, word distributions, presence of disfluencies and back-channelings, and so on \cite{biber1991variation}. This makes language models pre-trained on web text not suitable for processing spoken languages.
Third, text does not reflect the rich set of features conveyed by oral languages. Speech carries not only phonetic information, but also non-verbal vocalizations (laughter, voice clicks, filler vocalization, etc), rhythm and intonation (prosody), and emotional markers. All of these features could help, not only with generating more expressive speech \cite{ren2020fastspeech,lancucki2021fastpitch}, but also with the semantic analysis of the content of the message \cite{cutler1997prosody,tran2017parsing}.

To combat these deficiencies, more recently there is increasing interest in exploring speech pre-training using large quantities of unlabeled speech data~\cite{chung2019unsupervised,schneider2019wav2vec,kharitonov2020data,baevski2020wav2vec,hsu2020hubert,liu2020mockingjay,ling2020decoar,tjandra2020unsupervised,hsu2021robust,hsu2021hubert}. However, most of the studies evaluate their models on discriminative tasks, such as ASR and those in the SUPERB benchmark~\cite{Yang2020superb}. 
To the best of our knowledge, generative spoken language modelling (GSLM)~\cite{Lakhotia2021} is the only prior work that evaluates prompted speech completion, a generative tasks that is similar to the text completion task in GPT-2~\cite{radford2019language}. To remove the reliance on text, GSLM exploits discovered units from self-supervised models to build a unit language model (uLM) and a unit-to-spectrogram (u2S) model. Speech completion can be achieved by first sampling a unit sequence from the uLM with a unit prompt inferred from a speech prompt, and then synthesizing the sampled sequence into speech with the u2S model.
Unfortunately, because those discovered units encode mostly phonetic information~\cite{polyak2021speech}, it suffers from the same prosodic information loss issue as text-based LMs. Therefore, when using that uLM for speech completion, it fails to continue with a coherent tone to the prompt.

In this paper, we introduce a prosody-aware generative spoken language model (pGSLM) that jointly models phonetic content and prosody, in order to leverage prosody for comprehension, and to generate speech coherent with the prompt, which is a precursor for building speech-based dialogue systems. In keeping with our aim of liberating NLP from its over-reliance on text, we follow GSLM and represent the phonetic content with self-supervised units discovered from raw audio. As for prosody, it is represented by the pattern of quantized fundamental frequency (F0) and duration. pGSLM is comprised of two separately trained components: an auto-regressive Multi-Stream Transformer Language Model (MS-TLM) that predicts the next phonetic and prosodic representation given the past ones, and a unit High-Fidelity Generative Adversarial Network (HiFi-GAN) adapted from~\citet{polyak2021speech} that converts the MS-TLM output into a waveform like a vocoder.
To evaluate the proposed model, we adopt metrics from~\cite{Lakhotia2021} for content evaluation, and devise a series of metrics for prosody evaluation. Experimental results demonstrate that 1) joint modeling of prosody improves phonetic content modeling, 2) pGSLM can generate speech continuation coherent with the prompt in term of the content and the prosody, and 3) proper choices of model and prosodic representation is crucial to synthesizing natural, coherent, and expressive speech.


\section{Related Work}
Our work is related to utilizing prosody for comprehension and predicting prosody for speech synthesis, which we discuss in the following sections.
\subsection{Improving Comprehension with Prosody}
Prosody, which is often characterized by the rhythm, intonation, and intensity of speech, carries useful information for comprehending speech in addition to the textual content~\cite{cutler1997prosody}. Prior studies have shown that including prosody information can improve the performance from text-only models on speech segmentation~\cite{shriberg2000prosody}, dialogue act classification~\cite{shriberg1998can,ward2000prosodic}, syntactic parsing~\cite{tran2017parsing}, speech–language pathology~\cite{cohen2019predicting}, ASR~\cite{ostendorfy2003prosody,shriberg2004prosody}, and language modeling~\cite{huang2007using,su2008exploiting,ward2012prosodic}.
These studies provide strong empirical evidences for the benefit of considering prosody in processing spoken languages, especially in the conversational scenarios. 

This work shares the same motivation, but differs from the prior work in two crucial aspects. First, this work utilizes discrete units discovered from a self-supervised model and hence does not require any textual supervision, making it applicable to both written and unwritten languages, while in the prior work prosody information is used alongside text. Second, our model can be regarded as the speech version of GPT, which does not require any task-specific labels and can be pre-trained on large quantities of unlabeled speech data. The ability to leverage more data is shown to be the key to achieve good performance in text pre-training.

\subsection{Prosody Prediction for Speech Synthesis}
The proposed pGSLM model can be re-purposed as a text-to-speech (TTS) model when the phonetic content (represented as a unit sequence) is given and the prosody is generated by the MS-TLM model. This is similar to FastSpeech~\cite{ren2020fastspeech} and FastPitch~\cite{lancucki2021fastpitch} TTS models, where prosodic features are predicted from text and speech are generated conditioning on both the text and the predicted prosodic features. As FastSpeech and FastPitch are designed to improve the inference-time efficiency from auto-regressive models like Tacotron~\cite{wang2017tacotron}, they predict prosodic features and spectrograms without introducing dependency between time steps. In other words, these models assume that the prosody features within an utterance are not correlated across time steps given the text, whereas our proposed MS-TLM does not make such an assumption. We will demonstrate empirically the conditional independence is not a realistic assumption and our model achieves better performance on prosody metrics with auto-regressive modeling. 

As for analysis on prosody modeling, we present more extensive metrics by considering both teacher-forcing decoding and sampling, while prior work does not consider the multi-modal nature of prosody and only generate prosody deterministically~\citep{ren2020fastspeech}. Moreover, we also evaluate prosody in a more disentangled manner by measuring the error of the prosody prediction module alone instead of measuring the error of the prosody extracted from the synthesized waveform: the latter conflates the impact from both the prosody prediction module and the vocoder. 

\section{Method}\label{sec:methods}
In this section, we first describe the phonetic and prosodic representations used in pGSLM, and then introduce the two components it is comprised of: a multi-stream transformer language model and an adapted unit HiFi-GAN.

\subsection{Phonetic and Prosodic Representations}
\label{sec:representations}
We choose units with a vocabulary size of 100 derived from HuBERT~\cite{hsu2021hubert}, a self-supervised speech model, as the phonetic representation. Specifically, these units are obtained through clustering the 6th transformer layer output of the base HuBERT model provided in~\cite{hsu2021hubert} using a k-means algorithm, following the recipe of HuBERT closely. A speech waveform can therefore be encoded into a sequence of discrete units at a frame rate of 50 units per second, or alternatively, into a sequence of (unit, duration) tuples using run-length encoding.
HuBERT units were found to perform favorably compared to other self-supervised units such as wav2vec 2.0~\cite{baevski2020wav2vec} and VQ-VAE~\cite{van2017neural} in terms of lexical content modeling~\cite{Lakhotia2021} and disentangling prosodic information~\cite{polyak2021speech}.

We use \textit{unit duration} $d$ and \textit{fundamental frequency} (F0, or pitch) $f$ to derive prosodic representations. \citet{polyak2021speech} has shown that pairing HuBERT units with duration and F0 enables high-quality speech re-synthesis that preserves more prosodic information such as intonation compared to re-synthesizing with only units. Similar results are demonstrated in several other studies~\cite{ren2020fastspeech, lancucki2021fastpitch} in the context of text-to-speech synthesis.
Unfortunately, while F0 encodes prosodic information, it also encodes significant amount of speaker information. Figure~\ref{fig:emov_pitch} in the appendix illustrates how speaker and prosodic information (emotion) are disentangled in raw pitch using a multi-speaker multi-emotion dataset, EmoV~\cite{adigwe2018emotional}. We do not wish to model speaker variation in pGSLM because it is less relevant to spoken language understanding compared to prosody. To that end, we propose to model \textit{speaker-mean normalized log F0}: $\lf = \log f - \mathbb{E}_{f' \text{ from the same speaker as } f} [\log f']$,
which can be interpreted as the ratio to the mean pitch in the log space: $\lf = \log ( f / \bar{f} )$, where $\bar{f} = \exp \mathbb{E}_{f'} [\log f']$. Specifically, the equation above is used for voiced frames, and the expectation is taken over voiced frames from a speaker. For unvoiced frames, we simply set $\lf = 0$.

One may ask why F0 is only normalized by the speaker mean but not the variance. We argue that the variance encodes the ``level of expressiveness'' and it is desired to preserve it. This is demonstrated empirically in Figure~\ref{fig:logf0_mean_var} in the appendix, where speakers from expressive datasets, EmoV and Blizzard 2013~\cite{blizzard}, exhibits larger speaker log F0 standard deviation than those in less expressive datasets, LJSpeech~\cite{ljspeech17} and VCTK~\cite{veaux2016superseded}.
On the other hand, we also found that variance is more correlated mean in the linear space than in the log space, as shown in Figure~\ref{fig:f0_mean_var}. Therefore, we argue that mean-normalized log F0 is a more suitable representation for prosody as it encodes less speaker information while preserving the level of expressiveness. 

\subsection{Multi-Stream Transformer LM}
We adapt the Transformer LM from \cite{Lakhotia2021} to take multiple streams of input and predict multiple streams of output, and refer to it as the Multi-Stream Transformer Language Model (MS-TLM). An MS-TLM predicts a sequence of segment representations, which reduces the sequence length significantly and is found beneficial compared to predicting frame sequences~\cite{Lakhotia2021}. Each segment is represented with the unit $u$, duration (in frames) $d$, and normalized pitch $\lf$. The first two are obtained by run-length encoding the fixed frame rate unit sequence, while a segment $\lf$ is computed by averaging those from voiced frames within a segment or set to 0 if the entire segment is unvoiced. An example is provide in Appendix~\ref{sec:frm_to_seg}.

\subsubsection{Delayed prosody prediction}
\begin{figure}[ht]
    \centering
    \centering
    \includegraphics[width=\linewidth]{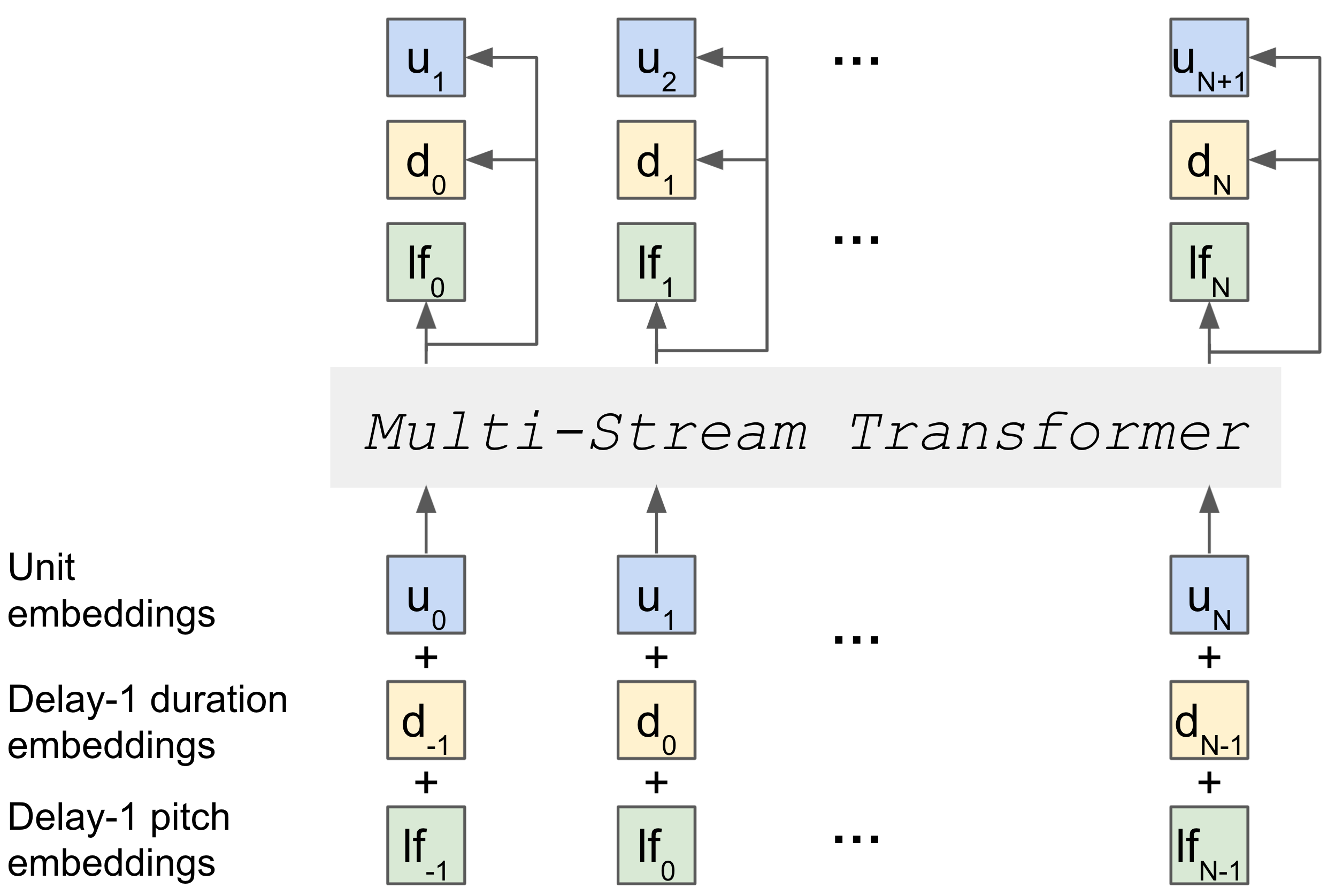}
    \caption{Delayed multi-stream transformer language model with prosody stream delay $\Delta=1$.}
    \label{fig:mstlm}
\end{figure}

Let subscript $t$ be the segment index. At each step, a vanilla MS-TLM takes $(u_{t-1}, d_{t-1}, \lf_{t-1})$ as input, linearly projects each of them to the dimension of the transformer, and feeds the summed embeddings to the transformer. The transformer output at that step is projected to the dimension of each stream to predict $u_t$, $d_t$, and $\lf_t$ independently. The distribution modeled by the synchronous MS-TLM $p(u_{1:T}, d_{1:T}, \lf_{1:T})$ can be written as:
\begin{align}
    \scriptstyle
    \prod_{t=1}^{T}\ & p(u_t \mid u_{1:t-1}, d_{1:t-1}, \lf_{1:t-1}) \nonumber \\ 
    \times\ & p(d_t \mid u_{1:t-1}, d_{1:t-1}, \lf_{1:t-1}) \nonumber \\
    \times\ & p(\lf_t \mid u_{1:t-1}, d_{1:t-1}, \lf_{1:t-1}).
\end{align}
We see that the factorial assumption here may be too strong, because the duration and the pitch of a segment are highly correlated with the phonetic content of the same segment. To alleviate that without introducing intra-step dependency or interleaving streams (which increases the sequence length and requires determining an order for the three streams a priori), we introduce a delay factor $\Delta$ ($\Delta \ge 0$) for prosodic streams, which shift prosodic input and output streams backward by $\Delta$ steps, taking $(u_{t-1}, d_{t-\Delta-1}, \lf_{t-\Delta-1})$ as input and outputting  $(u_t, d_{t-\Delta}, \lf_{t-\Delta})$. When $\Delta=1$, each step of the LM predicts the unit of the current segment and the prosodic representations of the previous segment, of which the lexical unit has been observed already, as shown in Figure~\ref{fig:mstlm}. 

\subsubsection{Quantizing prosodic representations}
\label{sec:quant_prosody}
A straightforward solution to encode prosody streams $d$ and $\lf$ is to represent them as continuous values and minimize an L1 or L2 loss for training, similar to FastSpeech2~\cite{ren2020fastspeech} and FastPitch~\cite{lancucki2021fastpitch}. Doing so assumes that the duration and the pitch of a segment follow a unimodal distribution (Laplace for L1 and Gaussian for L2) given the context. If the underlying distribution is multimodal with wide spread, the learned distribution would be significantly underfitting with a mean far from the modes.
Empirically, we found that such modeling indeed leads to predicting $\lf$ values very close to 0 for all segments, and the generated prosody sounds dull and boring.

Inspired by WaveNet~\cite{oord2016wavenet}, we represent prosodic features as discrete random variables through quantization. It is straightforward to quantize $d$ since it encodes integer values originally (length in frames). We set the maximum length to be 32 and the bin width to be 1, resulting in 32 bins. We quantize speaker-mean normalized log F0 $\lf$ into $K=32$ bins such that each bin with boundaries $[b_{i-1}, b_{i}]$ contains the same probability mass: $\mathbb{P}(\lf \in [b_{i-1}, b_{i}]) = 1/K$. 

\subsubsection{Training objective}
\label{ss:train}
The training loss is a weighted sum of three per-stream losses. Omitting dependency on the context for brevity, MS-TLM defines a distribution $p(u_t, d_t, lf_t)$ of the potential values for a timestep $t$. Then, denoting ground-truth per-channel values as $u_t^*, d_t^*, lf_t^*$, we get:
\begin{align}
    L&(p(u_t, d_t, lf_t), u_t^*, d_t^*, lf_t^*) = L_u( p(u_t), u_t^*) \nonumber \\
    & + \alpha \cdot L_d(p(d_t), d_t^*) + \beta \cdot L_{lf}(p(lf_t), lf_t^*)
\end{align}
In all experiments, we use cross-entropy as the loss on the predictions of the unit channel ($L_u$). Whenever we operate on quantized prosody values (both duration and F0), we also use cross-entropy as losses $L_d$ and $L_{lf}$. In the case of continuous-valued prosody streams, we treat predicted values $p(d_t)$ and $p(lf_t)$ as the mode of Laplacian distributions and maximize the log likelihood of the model, which is equivalent to minimizing an L1 loss.
In preliminary experiments, we found that the results are relatively robust to variations of the relative weights $\alpha$ and $\beta$, hence we fix them $\alpha = \beta = 0.5$ in all our experiments.

\subsubsection{Sampling from a model}
\label{ss:sampling}
To generate new utterances, potentially conditioned on a prompt, we run autoregressive generation where at each step we sample units, duration, and normalized log F0 values, append them to the context and feed them back. In the case of discrete channels (units, also duration/pitch in the case of discrete-valued models), we sample from the corresponding multinomial distribution. As commonly done in language modelling~\cite{Lakhotia2021}, we perform sampling with temperature by scaling the logits by the temperature parameter. We fine-tune the temperature on the validation data.

For MS-TLM that models normalized log F0 as continuous variables, we draw samples from a Laplacian distribution with its location parameter set to the predicted value, because the model assumes the output distribution is Laplacian (see \S-\ref{ss:train}). For duration, to avoid sampling invalid values, we sample from a Laplacian distribution truncated at zero and round it to the nearest positive integer.


\subsection{Waveform Generation with Unit Hifi-GAN}
Given $(u_{1:T}, d_{1:T}, \lf_{1:T})$ generated from the MS-TLM, we adapt the discrete unit-based HiFi-GAN vocoder from~\citep{polyak2021speech} to generate waveform. The original vocoder proposed in~\citep{polyak2021speech} takes in frame-level discrete unit, pitch and speaker embedding as input and applies VQ-VAE quantization on the pitch. As MS-TLM predicts quantized speaker-mean normalized log F0 on the segment level, we modify the training of the vocoder so that it takes frame-level segment-average pitch as input, where the pitch values for frames within a segment are set to the same value. We apply the same quantization described in \S~\ref{sec:quant_prosody} instead of VQ-VAE on the pitch. The unit Hifi-GAN and the MS-TLM are trained separately.
\section{Experimental Setup}
\subsection{Data, Model, and Training}
In our experiments, we train MS-TLM models on two English datasets: LibriSpeech~\cite{panayotov2015librispeech} and a 6K-hour subset~\cite{Riviere2020} of Libri-Light~\cite{kahn2020libri} which we refer to as LL-6K. Both datasets represent audio books and we use LibriSpeech dev-clean and test-clean as validation and test sets. As described in Section~\ref{sec:representations}, we use HuBERT-based unit representations.  However, to investigate whether our proposed models can work with other types of units, we also experiment with CPC~\cite{Riviere2020,oord2018representation} and ground-truth phone representations. We experiment with a vocabulary of 100 units when working with Hubert and CPC, following the same protocol and using the same pre-trained models as~\citet{Lakhotia2021}. On the other hand, frame-level phone transcripts are obtained through forced-alignment using the \texttt{tri6b} model from Kaldi's LibriSpeech recipe~\cite{povey2011kaldi}. The position- and context-independent phones without lexical stress markers are used, which include 41 units (39 phones, one silence \texttt{SIL}, and one spoken noise \texttt{SPN}). The frame rate of CPC and phone units is 100Hz, and is 50Hz for HuBERT units.

We experiment with MS-TLM of two sizes: base and large. The base one has 6 layers, 8 attention heads per layer, embedding size of 512. Its FFN layer has 2048 units. The large variant has 12 layers, each with 16 heads, embedding size of 1024 and the FFN layer is of dimensionality 4096. We set attention dropout and dropout probabilities to 0.1 for both alternatives. On top of that, we apply sequence-level and span-level~\cite{baevski2020wav2vec} input dropout to the two prosody streams. Specifically, each stream is zero-ed out with a probability of 0.2, and 2\% of the steps are selected as starts, from which 5 steps of that stream is zero-ed out.
Optimization is done using Adam~\cite{kingma2014adam} with a peak learning rate of~5e-4. Learning rate ramps up linearly for the first 4K updates, and then decays to 0 with an inverse square-root schedule. We train the base model for 70 epochs, and large model for 100 epochs. Each GPU's batch contains up to 3072 $(u, d, \lf)$ segments and we used 8 (16) GPUs to train base (large) MS-TLM. For each update, we aggregated gradients from 8 batches.

\subsection{Prosody and Content Evaluation}
Our overall goal is to find models that can freely generate meaningful content and consistent as well as diverse prosody. In this Section, we define a set of metrics that measure models' performance over each stream individually and combined, in both the teacher-forcing mode and the inference mode.

\subsubsection{Teacher-forcing metrics} 
A simple way to evaluate models is to measure its loss on hold-out data in a setup where for each step the full ground truth context is provided.
For the unit stream, we measures Negative Log-Likelihood (NLL), equivalent to cross-entropy. For the duration and pitch streams we use Mean Absolute Error (MAE), equivalent to L1 loss. When the pitch values are quantized, we de-quantize predictions to the means of the respective buckets.

\begin{figure}[t]
    \centering
    \includegraphics[trim=190 50 90 140,clip,scale=0.5]{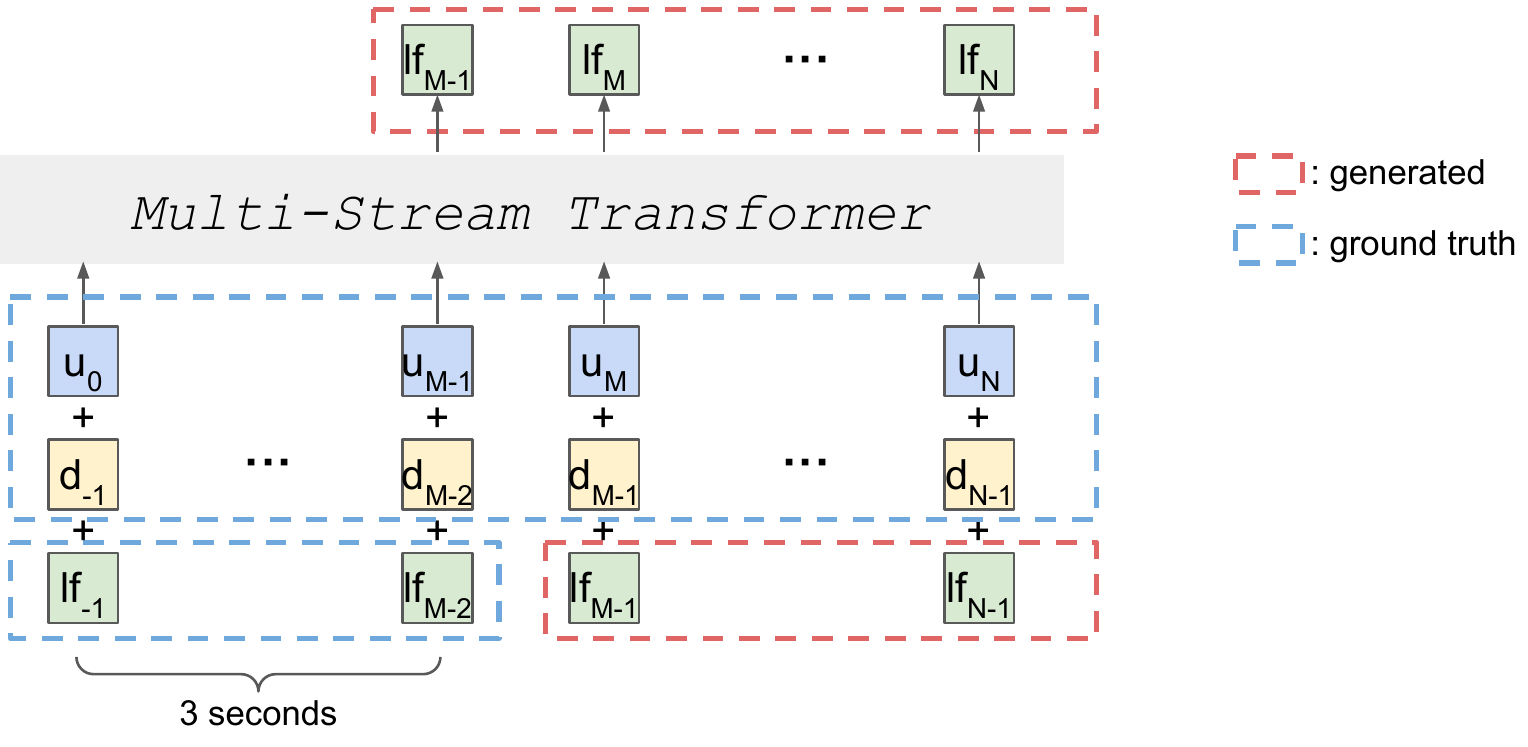}
    \caption{Per-stream prosody continuation task. $lf$ is the target stream for continuation in this example.}
    \label{fig:psd_cont}
    
    \centering
    \includegraphics[trim=190 75 90 60,clip,width=\linewidth]{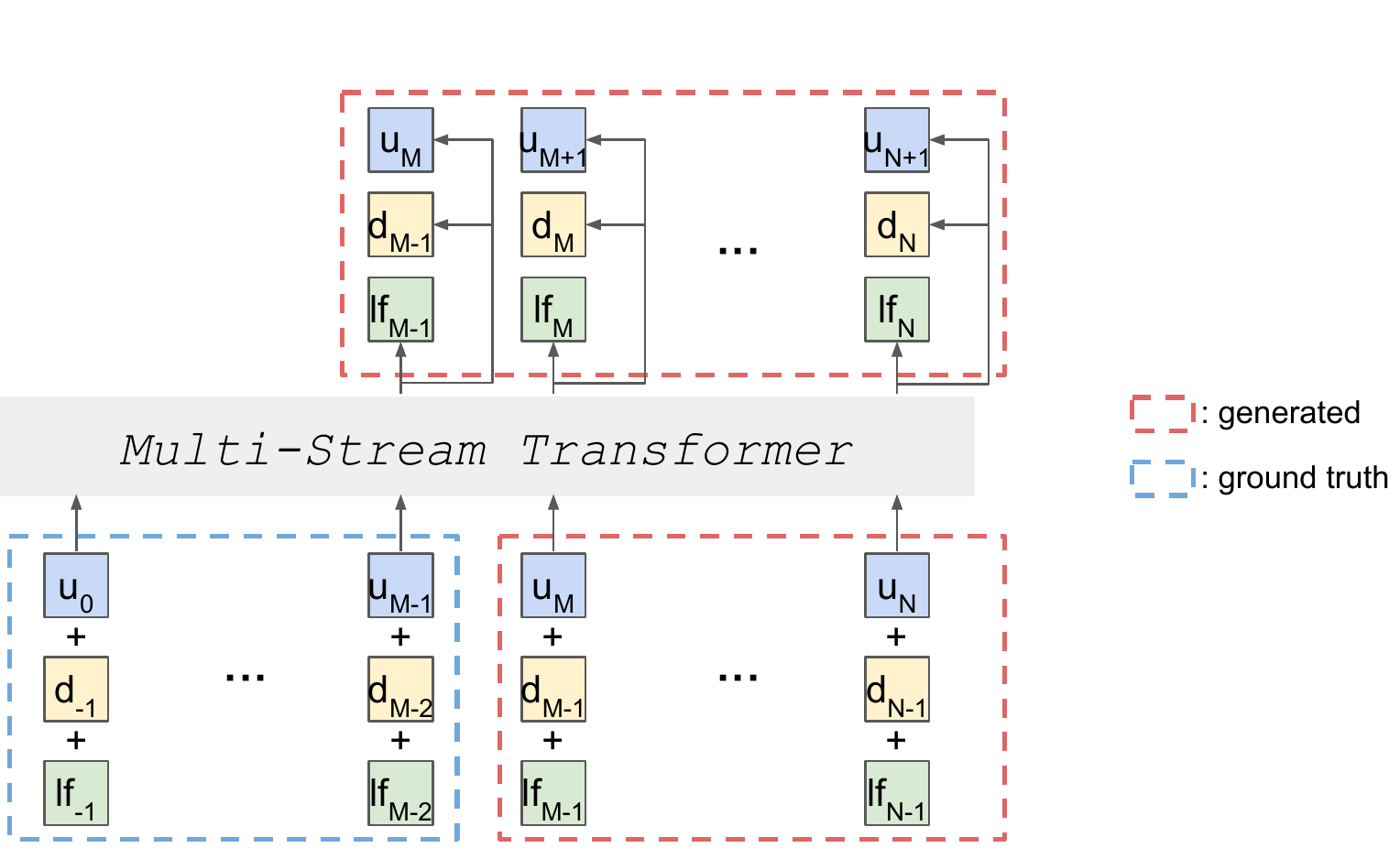}
    \caption{Speech continuation task.}
    \label{fig:sp_cont}
\end{figure}

\subsubsection{Per-stream prosody continuation}\label{sec:per_stream}
We next evaluate the model's ability to complete a stream in isolation. Specifically, we provide a 3s prompt for all streams, and then sample auto-regressively the target stream while feeding the ground truth value for the other streams, as depicted in Figure~\ref{fig:psd_cont}. The prompts are inferred from the utterances in the validation set. When prosodic features are quantized, we sample with a temperature $\tau \in \{0.0, 0.25, 0.5, 0.7, 1.0, 1.3\}$, and when they are continuous, we sample with a scale $b \in \{0.0, 0.05, 0.125, 0.25, 0.5, 0.7, 1.0, 1.3\}$ for duration and 
$b \in 0.01 \times \{2^{-6}, 2^{-5}, \cdots, 2^{0}\}$ for pitch. The temperature/scale is chosen to minimize the Min-MAE for the corresponding stream, which we describe next. We chose different sweeping ranges for continuous pitch and duration because they have different inherent standard deviations.

\paragraph{Correctness (Min-MAE)}
A prompt might have multiple meaningful continuations in the content space~\cite{Lakhotia2021}. Similarly, a single sentence can have multiple correct prosodic profiles. To account for that, for each prompt we generate $n=20$ samples so that a model has a chance to cover most modes of the underlying distribution, and report the minimal MAE (min-MAE) against the reference among the $n$ samples. 

\paragraph{Consistency (Corr.)} To quantify the models' capability to generate consistent prosody, we measure Pearson correlation between the mean values of a stream in the prompt and in the generated continuation. Clearly, if the prompt has a distinct tempo or a pitch, a good continuation should reflect this. The same setup as the min-MAE metric is used ($n=20$) with one exception: we only consider sequences that are at least 6s long.

\paragraph{Expressiveness (Std.)} To measure how expressive the generated prosody is, we calculate the standard deviation of the generated values and expect a good model to exhibit a similar level of that as the ground truth. The same setup as in ``Min-MAE'' is used.

\subsubsection{Speech continuation}
Lastly, we evaluate the model's ability to carry out prompted speech completion, where all three streams are sampled given a 3s prompt using the temperature/scale parameter determined from per-stream continuation (\S~\ref{sec:per_stream}) as illustrated in Figure~\ref{fig:sp_cont}. We sample the MS-TLM auto-regressively until it emits the EOS unit or reaches the length of the reference. The MS-TLM output is synthesized into a waveform using the adapted HiFi-GAN.

\paragraph{Content (Max-Word-Cont-BLEU2)} We re-use the maximum word-level continuation BLEU2 proposed by \citet{Lakhotia2021} to quantify how well a model can complete a prompt in terms of the textual content. We transcribe the waveform with an off-the-shelf wav2vec 2.0-based ASR~\cite{baevski2020wav2vec} (same as~\cite{Lakhotia2021}) and compute the BLEU2 score for each of the $n=20$ continuations against the reference completion. The highest one is used as score for a prompt.

\paragraph{Human evaluation (MOS, MMOS, PMOS)}
We ask humans to evaluate three aspects of speech continuation: sound quality, meaningfulness (how natural the text content is considering both grammar and meaning), and prosody (how consistent and natural the intonation and the rhythm is). We follow the human evaluation protocol used by \citet{Lakhotia2021} closely, where raters evaluate subjective quality of the recordings using headphones on a scale between 1 to 5 with an increment of 1, the higher the better. Only Native English speakers were recruited as raters for all three studies. The same 100 prompts as \cite{Lakhotia2021} from LibriSpeech test-other are used, and each system generates one continuation per prompt. Each continuation is evaluated by at least 5 raters for each aspect. The CrowdMOS package~\cite{ribeiro2011crowdmos} was used for all experiments using the recommended recipes for outlier removal. All participants were recruited using the Amazon Mechanical Turk platform. The metrics on the three aspects are denoted as MOS, M-MOS, and P-MOS.






\section{Results}

\subsection{Prosodic Inputs Are Useful for Content and Prosody Modeling}\label{sec:contentmodel}
\begin{table}[h]
    \centering
    \resizebox{\linewidth}{!}{
    \begin{tabular}{ccccc|ccc}
        ID & Input & Output & Quant? & $\Delta$ & $u$ NLL$\downarrow$ & $d$ MAE$\downarrow$ & $\lf$ MAE$\downarrow$\\
        \midrule\midrule
        \multicolumn{8}{l}{\textbf{\textit{Base MS-TLM, HuBERT units, trained on LS960}}}\\
        1 & $u$ & $u$ & n/a & n/a & 1.522 & n/a & n/a \\
        2 & $u$ & $(u, d, lf)$ & \checkmark & 0 & 1.525 & 0.759 & 0.115 \\
        3 & $u$ & $(u, d, lf)$ & \checkmark & 1 & 1.517 & 0.586 & 0.112 \\
        4 & $u$ & $(u, d, lf)$ &            & 1 & 1.514 & 0.562 & 0.093 \\
        \cmidrule{2-8}
        5 & $(u, d, lf)$ & $u$          & \checkmark & 0 & \textbf{1.336} & n/a & n/a \\
        6 & $(u, d, lf)$ & $(u, d, lf)$ & \checkmark & 0 & 1.337 & 0.722 & 0.052  \\
        7 & $(u, d, lf)$ & $(u, d, lf)$ & \checkmark & 1 & 1.441 & 0.551 & 0.049 \\
        8 & $(u, d, lf)$ & $(u, d, lf)$ &            & 1 & 1.447 & \textbf{0.536} & \textbf{0.046} \\
        \midrule
        \multicolumn{8}{l}{\textbf{\textit{Large MS-TLM, HuBERT units, trained on LL6k}}}\\
        9  & $u$          & $(u, d, lf)$ &            & 1 & 1.513 & 0.563 & 0.095 \\
        10 & $u$          & $(u, d, lf)$ & \checkmark & 1 & 1.522 & 0.586 & 0.116 \\
        \cmidrule{2-8}
        11 & $(u, d, lf)$ & $(u, d, lf)$ &            & 1 & 1.421  & \textbf{0.527} & \textbf{0.043} \\
        12 & $(u, d, lf)$ & $(u, d, lf)$ & \checkmark & 1 & \textbf{1.406} & 0.543 & 0.047 \\
        \midrule
        \multicolumn{8}{l}{\textbf{\textit{Base MS-TLM, CPC units, trained on LS960}}}\\
        13 & $u$          & $(u, d, lf)$ & \checkmark & 1 & 1.511 &  1.302 & 0.122 \\
        14 & $(u, d, lf)$ & $(u, d, lf)$ & \checkmark & 1 & \textbf{1.353}  & \textbf{1.181} & \textbf{0.045} \\
        \midrule
        \multicolumn{8}{l}{\textbf{\textit{Base MS-TLM, Phone units, trained on LS960}}}\\
        15 & $u$          & $(u, d, lf)$ & \checkmark & 1 & 1.559 & 2.748 & 0.150 \\
        16 & $(u, d, lf)$ & $(u, d, lf)$ & \checkmark & 1 & \textbf{1.485} & \textbf{2.419} & \textbf{0.079}\\
        
    \end{tabular}
    }
    
    \vspace{2em}
    
    \resizebox{\linewidth}{!}{
    \begin{tabular}{ccccc|ccc}
        \multirow{2}{*}{ID} & \multirow{2}{*}{Input} & \multirow{2}{*}{Output} & \multirow{2}{*}{Quant?} & \multirow{2}{*}{$\tau$} 
        & $u$ NLL & $d$ MAE & $\lf$ MAE\\
        & & & & & stddev & stddev & stddev \\
        \midrule\midrule
        \multicolumn{8}{l}{\textbf{\textit{Base MS-TLM, HuBERT units, trained on LS960}}}\\
        2 & $u$ & $(u, d, lf)$ & \checkmark & 0 & 0.0004 & 0.00226 & 0.0033 \\
        \cmidrule{2-8}
        6 & $(u, d, lf)$ & $(u, d, lf)$ & \checkmark & 0 & 0.0022 & 0.00306 & 0.00037 \\
    \end{tabular}
    }
    \caption{(Top) Teacher-forcing metrics on Librispeech dev-clean. Exp 1 is identical to the uLM presented in \cite{Lakhotia2021}. We can observe that models with both phonetic and prosodic input $(u, d, \lf)$ consistently outperforms their counterpart model with only phonetic input $u$. This trend holds for different lexical representations (HuBERT, CPC, phone), both continuous and discrete prosodic features, and different delay factors $\tau$. (Bottom) We train Exp ID 2 and 6 with five random seeds and measure the standard deviation on all three metrics. Results show that the gap between the models with and without prosodic input is significant relative to the standard deviation.
    \label{tab:tf_metrics}}
\end{table}


\begin{table*}[ht]
    \centering
    \resizebox{.9\linewidth}{!}{
    \begin{tabular}{cccc|ccc|ccc|c}
        ID & Input & Output & Quant? & \multicolumn{3}{c|}{$d$} & \multicolumn{3}{c|}{$\lf$} & Max-Word- \\
        & & & & min-MAE $\downarrow$ & Corr.$\uparrow$  & Std.$\uparrow$ & min-MAE $\downarrow$ & Corr.$\uparrow$ & Std.$\uparrow$ & Cont-BLEU2 $\uparrow$ \\
        \midrule\midrule
          & \multicolumn{2}{c}{ground truth} & n/a        & .000 & .463 & 1.32 & .000 & .520 &  .163 & 1.000 \\
          & \multicolumn{2}{c}{resynthesized}& \checkmark & .000 & .464 & 1.32 & .000 & .315 & .145  & .943 \\
        \midrule
        \multicolumn{11}{l}{\textbf{\textit{Large MS-TLM, HuBERT units, trained on LL6k, $\Delta=1$}}}\\
        9  & $u$          & $(u, d, lf)$      &            & .542 & .176 & .942 & .084 & .093 & .081 &  .488 \\
        10 & $u$          & $(u, d, lf)$      & \checkmark & .542 & .086 & \textbf{.965 }& .096 & .217 & .147 & .489 \\
        11 & $(u, d, lf)$ & $(u, d, lf)$      &            & .539 & \textbf{.344 }& .940 & .081 & \textbf{.494 }& .076 &  .498 \\
        12 & $(u, d, lf)$ & $(u, d, lf)$      & \checkmark & \textbf{.536} & .242 & .946 & \textbf{.077} & .324 & \textbf{.149 }& \textbf{.499} \\
    \end{tabular}
    }
    \caption{Continuation metrics on Librispeech test-clean. The temperature and the scale parameters used for sampling are selected using the dev-clean set. We compare the four large MS-TLM models from Table~\ref{tab:tf_metrics} here. For each utterance in the test set, we use the first 3 seconds as the prompt and sample 20 continuations.}
    \label{tab:continuation}
\end{table*}


\begin{table}[ht]
    \centering
    \resizebox{\linewidth}{!}{
    \begin{tabular}{cccc|ccc}
        ID & Input & Output & Quant? & \multicolumn{3}{c}{Mean Opinion Score} \\
        & & & & MOS & M-MOS & P-MOS \\
        \midrule\midrule
           & \multicolumn{2}{c}{resynthesized} & & 3.21$\pm$0.09 & 3.95$\pm$0.32 & 3.87$\pm$0.45 \\
        \midrule
        \multicolumn{7}{l}{\textbf{\textit{Large MS-TLM, HuBERT units, trained on LL6k, $\Delta=1$}}}\\
        9  & $u$          & $(u, d, lf)$      &            & 3.16$\pm$0.19 & 3.80$\pm$0.25 & 3.69$\pm$0.42 \\
        10 & $u$          & $(u, d, lf)$      & \checkmark & 2.66$\pm$0.18 & 3.36$\pm$0.40 & 3.15$\pm$0.52 \\
        11 & $(u, d, lf)$ & $(u, d, lf)$      &            & 3.31$\pm$0.23 & 3.76$\pm$0.27 & \textbf{3.78$\pm$0.46} \\
        12 & $(u, d, lf)$ & $(u, d, lf)$      & \checkmark & \textbf{3.43$\pm$0.20} & \textbf{4.04$\pm$0.20} & 3.75$\pm$0.48 \\
    \end{tabular}
    }
    \caption{Human evaluation on sound quality (MOS), meaningfulness (M-MOS), and prosody (P-MOS). $\pm$~indicates 95\% CI.}
    \label{tab:mos}
\end{table}

In Table~\ref{tab:tf_metrics} we report teacher-forcing metric calculated on LibriSpeech dev-clean dataset for a diverse set of models. In rows 1-8, we report metric values for base MS-TLM models that are trained on LibriSpeech 960h transcribed into HuBERT-100 units. In rows 9-12 we consider large MS-TLM models trained on HuBERT transcripts of LL6k. Rows 13 \& 14 and 15 \& 16 contain metric values for models that are trained on LibriSpeech 960h transcribed using CPC and ground-truth phonetic units.\footnote{Note: the metric values in this section are only comparable within the same unit type. To compare across unit types, one can synthesize the MS-TLM output into waveform and transcribe the speech with an ASR systems to compute metrics in the word or character space.} The row 1 corresponds to the prosody-ignorant baseline model of~\cite{Lakhotia2021}.

On comparing two models that only predict units (rows 1 and 5) we see that by simply adding prosodic channels to the input of the model, we obtain considerably lower level of negative log-likelihood of the units ($u$ NLL: 1.522 vs.\ 1.336). The same trend persist for the models that predict prosodic channels, too. For instance, this holds in the case of the continuous-F0 models (rows 9 \& 11: 1.513 vs.\ 1.421) and, equally for the quantized F0 HuBERT-based models (rows 10 and 12: 1.522 vs.\ 1.406). Moreover, this holds for the CPC-based models (row 13 \& row 14) and even for the models trained on phone transcripts (rows 15 \& 16). Hence we conclude that prosodic input \textit{universally} improves speech ``content'' modelling.

Our results in Table~\ref{tab:tf_metrics} also allow us to investigate whether shifting prosody streams w.r.t.\ the unit stream ($\Delta > 0$) is useful. On comparing rows 6 \& 7 we see that this is indeed the case: at an expense of some increase in $u$ NLL (e.g., $1.337$ vs.\ $1.441$) we obtain considerable relative improvement in $d$ MAE ($0.722 \rightarrow 0.551$). The trend follows when further increasing $\Delta$.
We also observe that having prosody in the context is beneficial when modeling prosody itself. Indeed, this is the case across all pairs of models (rows 9 \& 11, 10 \& 12) according to $d$ MAE and $lf$ MAE metrics. Moreover, this holds for the types of units that differ from HuBERT (CPC: rows 13 \& 14, phonetic units: rows 15 \& 16).


\subsection{Prosodic Inputs Are Useful for Speech Generation}
\begin{figure}[ht]
    \centering
    \includegraphics[trim=45 0 60 0, clip, width=\linewidth]{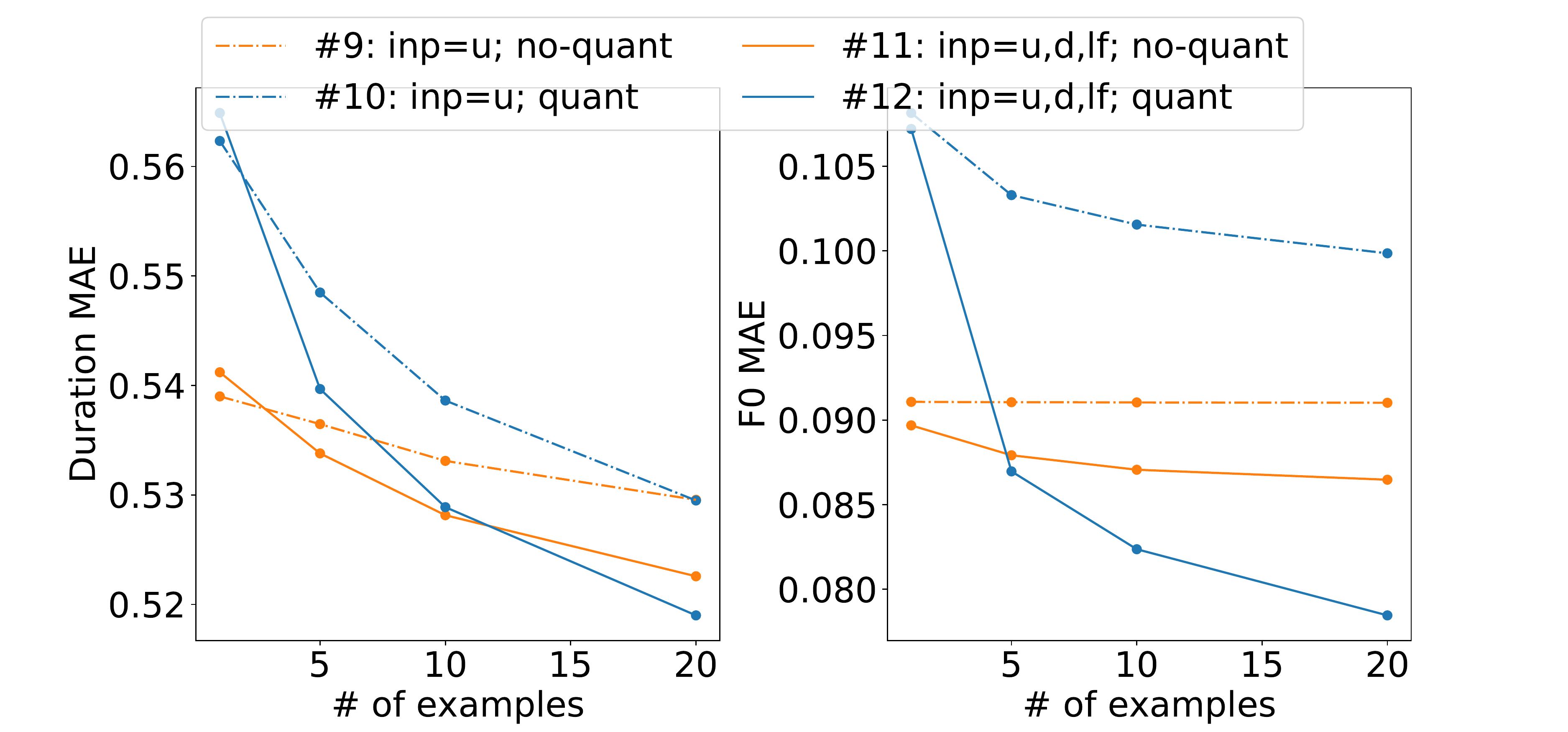}
    \caption{Minimal-MAE of duration (left) and pitch (right) with respect to different number of samples for the four large models (ID 9-12).}
    \label{fig:sampling}
\end{figure}

In our next experiment we study how the number of sampled prompt continuation affects prosody accuracy metrics (MAE). We report results for the four large models (rows 9-12) in Figure~\ref{fig:sampling}. From these results we observe that models that operate on quantized prosodic streams greatly benefit from sampling multiple candidates. In contrast, the two continuous-valued models seem to benefit little if at all (in the case of the F0 stream). We hypothesise that this striking difference is due to the ability of the multinomial-sampled trajectories to cover multiple mode of the underlying distribution, while the continuous-valued models produce samples that are ``averaged'' to the median of the underlying distribution due to the L1 loss.

In Table~\ref{tab:continuation} we report the continuation metrics for four large MS-TLM models, trained on HuBERT transcripts of LL-6k (they correspond to rows 9-12 in Table~\ref{tab:tf_metrics}).\footnote{Audios samples of speech continuation are included in the supplementary material.} These models differ in whether they have prosodic input or not (rows 11 \& 12 vs.\ 9 \& 10) and if the prosodic channels are discretized or not (10 \& 12 vs.\ 9 \& 11).

Firstly, on comparing models with and without prosodic input, we observe that having prosody in input improves the accuracy of the prosody continuation (in terms of MAE). This holds for predicting duration (e.g., 0.542 and 0.536 for rows 10 and 12). We see a higher relative difference for $lf$ (e.g., 0.096 vs.\ 0.077, same models). Our proposed models are also able to leverage provided prosody input to maintain high consistency of the prosody continuation, as measured by the correlation metrics. For example, for the continuous-prosody models the correlation values grows from 0.176 to 0.344 for the duration prediction and from 0.093 to 0.494 for the F0 channel. Having prosody input also turns out to be important for the word-level BLEU metric: models 11 and 12 outperform their counterparts without prosody inputs, 9 and 10.

Next, when contrasting discrete- and continuous-prosody models the following picture emerges. For both duration and F0 channels, discrete models achieve lower min-MAE errors. Further, both discrete models generate considerably more diverse F0 values than either of the continuous models (up to 2x higher std). Among the models with prosody inputs, the one with discrete prosody get higher variability in the $d$ channel. In contrast, the correlation metrics favor the prosody-aware continuous model. From the point of view of the word-level BLEU scores, both models are very close with the quantized model (row 12) being slightly ahead.  We attribute this difference between the models to the ability of discrete-valued MS-TLM to better describe multi-modal distributions, as we saw above in the experiment reported in Figure~\ref{fig:sampling}.

Table~\ref{tab:mos} presents the human evaluation results. The model with prosody input and quantized prosody performs significantly better than the rest on MOS and M-MOS, and is on par with the variant with prosody input and continuous prosody on P-MOS. Note that when not having the prosody input, the model with quantized prosody performs significantly worse on all metrics, demonstrating the importance of auto-regressive generation for discrete representation.

To summarize, we conclude that (i) including prosody input allows better modelling of speech, and (ii) architectures that operate with quantized prosody values, generally, perform better on our introduced metrics.

\section{Conclusion and Future Work}
In this work, we propose a text-free prosody-aware generative spoken language model, pGSLM, which models textual content and prosodic information explicitly and does not use any text supervision by leveraging self-supervised units. Through extensive evaluation on a diverse set of metrics, we demonstrated that prosody not only improves content modeling, but also enables better prompted speech generation that is aware of both the content and the prosody from the prompt for the first time in the literature. We conducted a number of ablation studies to validate the effectiveness of model design choices.

As for broader impacts, this work serves as the foundation for building better conditional speech generation applications where prosody is essential, such as in the conversational scenarios. In addition, the proposed model could also serve as a pre-trained model for other classification tasks, such as emotion recognition or syntactic parsing from speech, or as a pre-trained model for generative tasks such as text-to-speech synthesis with more expressive and coherent prosody. Finally, the proposed prosody metrics (teacher-forcing duration and pitch MAE, continuation correctness/consistency/expressiveness) may also be used for evaluation of text-to-speech synthesis systems that can produce diverse prosody for a given text input.


\clearpage
\bibliography{main}
\bibliographystyle{acl_natbib}

\clearpage
\counterwithin{figure}{section}
\appendix
\section{Analysis of Log F0 Distribution}
\label{sec:appendix}

\begin{figure}[ht]
    \centering
    \includegraphics[width=\linewidth]{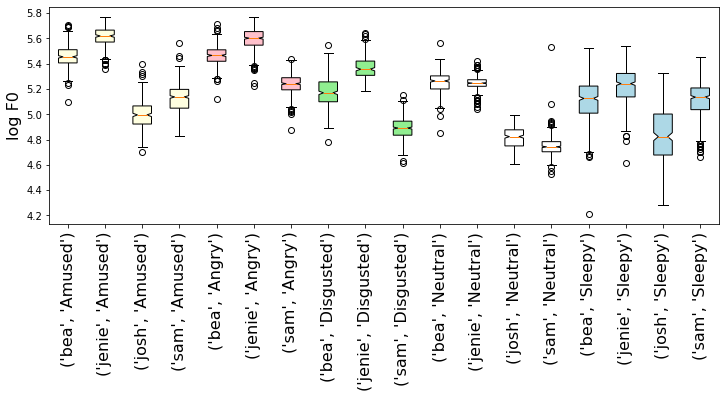}
    \includegraphics[width=\linewidth]{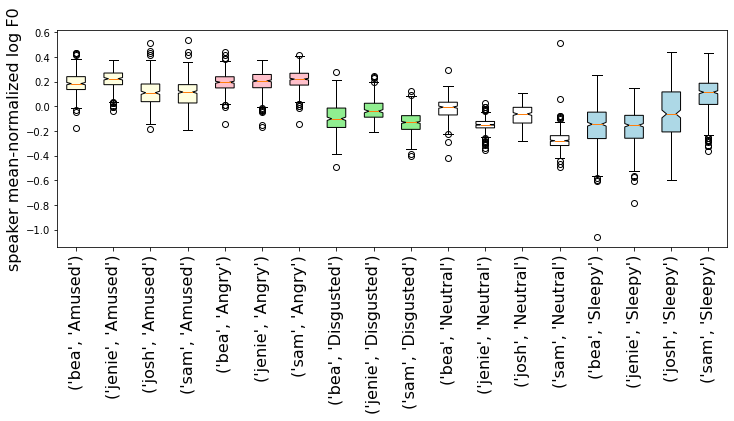}
    \caption{Log F0 distribution of each (speaker, emotion) combination in the EmoV dataset without speaker mean normalization (top) and with speaker normalization (bottom). Each color corresponds to one emotion.}
    \label{fig:emov_pitch}
\end{figure}

\begin{figure}[ht]
    \centering
    \includegraphics[width=\linewidth]{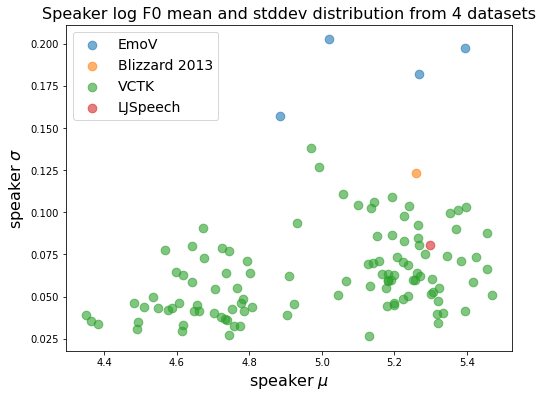}
    \caption{Speaker log F0 mean and standard deviation distributions from two expressive datasets (EmoV and Blizzard 2013) and two plain datasets (LJSpeech and VCTK). Each point corresponds to one speaker.}
    \label{fig:logf0_mean_var}
\end{figure}

\begin{figure}[ht]
    \centering
    \includegraphics[width=\linewidth]{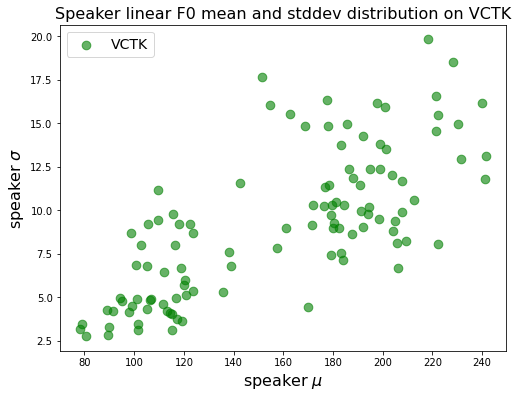}
    \caption{Speaker linear F0 mean and standard deviation distributions on VCTK. Each point corresponds to one speaker.}
    \label{fig:f0_mean_var}
\end{figure}


\section{HiFi-GAN Adaptation Analysis}
Table~\ref{tab:hifi_gan_adapt} presents an analysis of HiFi-GAN performance when using different quantized pitch representations. Similarly to~\cite{polyak2020unsupervised} we report voice decision error  (VDE)~\cite{nakatani2008method}, which measures the portion of frames with voicing decision error and F0 Frame Error (FFE)~\cite{chu2009reducing}, which measures the portion of frames that contain a deviation of more than 20\% in pitch value or have a voicing decision error. Results show that the chosen quantizer achieve favorable performance in terms of VDE and comparable results in terms of FFE without having to pre-train a F0 VQ-VAE quantizer.

\begin{table}[tbh]
    \centering
    \resizebox{\linewidth}{!}{
    \begin{tabular}{cccc|cc}
        \multicolumn{4}{c|}{F0} 
        & \multirow{2}{*}{FFE$\downarrow$} 
        & \multirow{2}{*}{VDE$\downarrow$} \\
        scale & norm. & res. & quant.  \\
        \midrule\midrule
        \rowcolor{black!10!white} \multicolumn{6}{c}{\cite{polyak2021speech}}\\
        lin & mean+std & frm & VQ-VAE (V) & 0.223 & 0.169 \\
        \midrule
        lin & mean+std & frm & VQ-VAE (B) & 0.198 & 0.181 \\
        lin & mean+std & frm & naive & 0.172 & 0.138 \\
        lin & mean+std & seg & VQ-VAE (V) & 0.149 & \textbf{0.116} \\
        lin & mean & frm & VQ-VAE (V) & 0.220 & 0.178 \\
        log & mean+std & frm & VQ-VAE (V) & 0.388 & 0.188 \\
        \midrule
        \rowcolor{black!10!white} \multicolumn{6}{c}{(This work)}\\
        log & mean & seg & naive & \textbf{0.134} & 0.118 \\
    \end{tabular}
    }
    \caption{Speech resynthesis results on the Blizzard 2013 validation set with segment-level pitch information. All the HiFi-GAN models are trained on the Blizzard 2013 training set using the same HuBERT units but different quantized pitch representations. \textbf{``scale''} denotes the F0 scale, which is linear (\textit{lin}) or logarithmic (\textit{log}). \textbf{``norm.''} denotes the normalization method applied to F0, which is normalizing by \textit{mean} or by mean and standard deviation (\textit{mean+std}). \textbf{``res.''} denotes  the resolution of pitch in the training time, where ``\textit{frm}'' refers to using frame-level pitch and ``\textit{seg}'' refers to using segment-level pitch (pitch values are set to the average for all frames within a segment). \textbf{``quant.''} denotes the F0 quantizer, where \textit{VQ-VAE (V)} is a neural pitch quantizer pre-trained on VCTK, \textit{VQ-VAE (B)} is one pre-trained on Blizzard, and \textit{naive} is the one adopted in this work.}
    \label{tab:hifi_gan_adapt}
\end{table}

\section{Example of Converting Frame-Level to Segment-Level Representations}
\label{sec:frm_to_seg}
Assume we have an utterance of six frames: [(13, 1.5), (13, 2.5), (13, 0.0), (21, 0.0), (27, 1.3), (27, 3.5)] where the first number in each tuple denotes the unit of the frame and the second number denotes the speaker normalized log F0 of the frame. In particular, the third and the fourth frame are unvoiced and their $lf$ values are set to 0.0. 

The segment level representation of the utterance is [(13, 3, 2.0), (21, 1, 0.0), (27, 2, 2.4)]. The first segment (13, 3, 2.0) is labeled with unit $u=13$, duration $d=3$ frames, and an average normalized log F0 $lf=(1.5+2.5)/2=2.0$ for the two voiced frames. The second segment contains only one unvoiced frame, and hence $lf$ is set to 0. Finally, the last segment contains two voiced frames, and therefore $d=2$ and $lf = (1.3+3.5)/2 = 2.4$.

\section{Effects of F0 Representation on MS-TLM}
Table~\ref{tab:f0_repr_nll} compares content modeling performance when using different pitch representations. Results show that using mean normalized pitch information is better than using raw pitch, and using log pitch is better than using linear pitch.

\begin{table}[ht]
    \centering
    \begin{tabular}{cc|c}
        F0 scale & F0 norm. & $u$ NLL \\
        \midrule\midrule
        linear & none & 1.763 \\
        linear & mean & 1.564 \\
        log & none & 1.461 \\
        \midrule
        \rowcolor{black!10!white} \multicolumn{3}{c}{(This work)} \\
        log & mean & \textbf{1.447} \\
    \end{tabular}
    \caption{Content modeling performance of base MS-TLM models trained on LS960 with HuBERT units using different pitch representations without quantization. \textbf{``scale''} denotes the F0 scale, which is linear (\textit{lin}) or logarithmic (\textit{log}). \textbf{``norm.''} denotes the normalization method applied to F0, which is not normalizing (none) or normalizing by mean.}
    \label{tab:f0_repr_nll}
\end{table}

\section{More Details of Human Evaluation}
\begin{figure*}
    \centering
    \includegraphics[width=\linewidth]{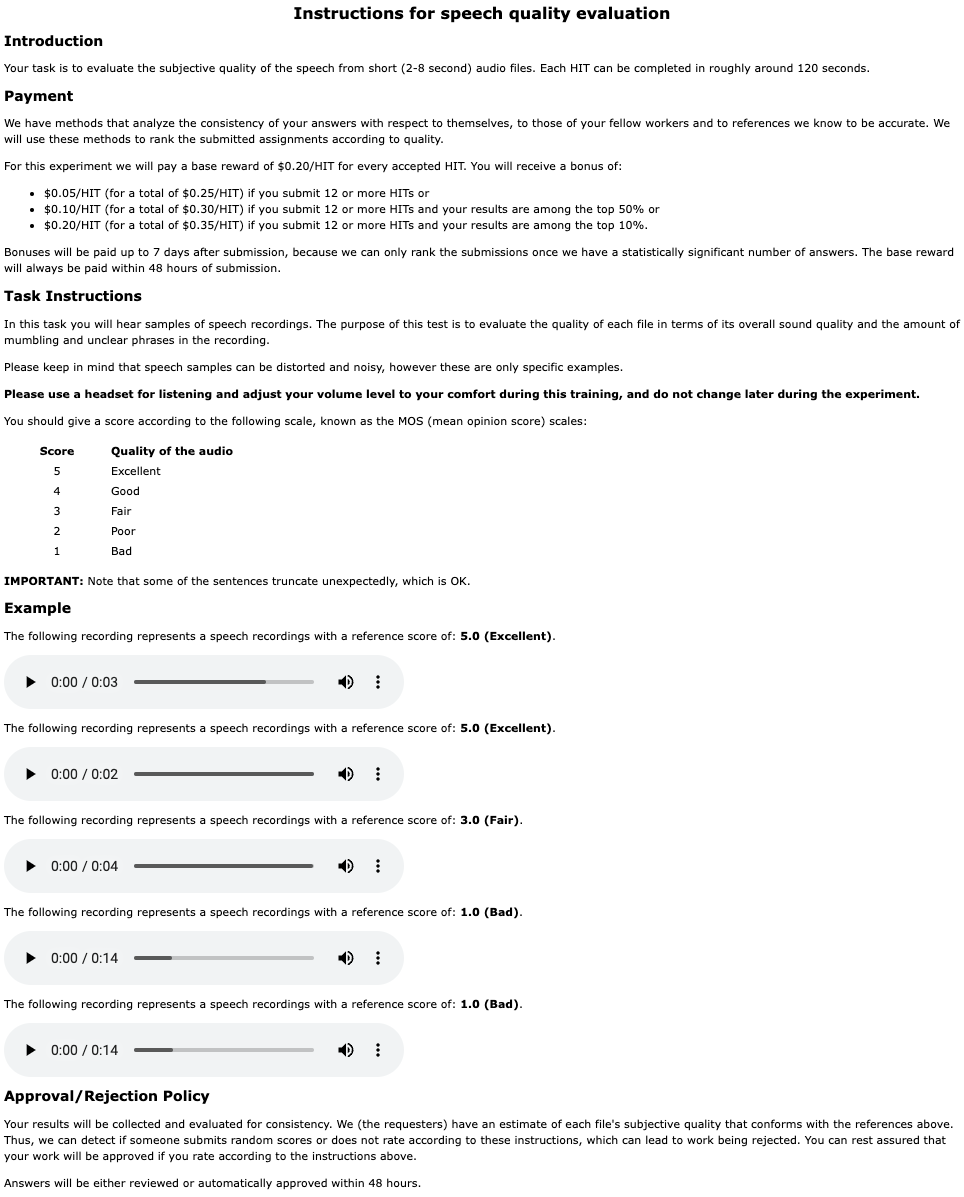}
    \caption{The instruction page for human evaluation on MOS.}
    \label{fig:mos_tmpl}
\end{figure*}

The instruction page displayed to the raters are shown in Figure~\ref{fig:mos_tmpl}. We modify the \textit{Introduction}, \textit{Task Instruction}, \textit{Example} in the instruction page for MOS, MMOS, and PMOS correspondingly. The text used for each metric are detailed in Table~\ref{tab:mos_text}

\begin{table*}[htb]
    \centering
    \begin{tabular}{l|p{6cm}|p{7cm}}
        Metric & Introduction & Metric-Specific Task Instruction \\
        \midrule\midrule
        MOS 
        & Your task is to evaluate the \textbf{subjective quality} of the speech from short (2-8 second) audio files. Each HIT can be completed in roughly around 120 seconds.
        & ...The CONTINUATION has been generated by a computer and your task will be to concentrate specifically on it and evaluate its \textbf{quality} in terms of the \textbf{sound clarity} on a 1 to 5 scale, (irrespective of the intonation or meaning) \\
        \midrule
        MMOS 
        & Your task is to evaluate the \textbf{subjective meaningfulness} of the speech from short (2-8 second) audio files. Each HIT can be completed in roughly around 120 seconds.
        & ...The CONTINUATION has been generated by a computer and your task will be to concentrate specifically on it and evaluate its \textbf{meaning} in terms of \textbf{grammar and content} on a 1 to 5 scale, (irrespective of sound clarity or intonation)\\
        \midrule
        PMOS 
        & Your task is to evaluate the \textbf{subjective prosodic coherence} of the speech from short (2-8 second) audio files. Each HIT can be completed in roughly around 120 seconds.
        & ...The CONTINUATION has been generated by a computer and your task will be to concentrate specifically on it and evaluate its \textbf{naturality} in terms of \textbf{intonation and rhythm} on a 1 to 5 scale, (irrespective of sound clarity or meaning) \\
    \end{tabular}
    \caption{The \textit{Introduction} and {Task Instruction} used for the three human evaluation metrics. All the \textit{Task Instruction} starts with ``In this task, you will hear samples of speech recordings, composed of the following parts: [PROMPT] beep [PROMPT CONTINUATION]. [PROMPT] is a sentence beginning, said by one voice. (“ANOTHER PREACHER AFTER REPROACHING HIM” in a male voice). Beep is a short tone. [PROMPT CONTINUATION] a longer sentence in a different voice, which starts with the initial prompt (“ANOTHER PREACHER AFTER REPROACHING HIM” in a female voice) and adds a few more words that continue this prompt (“TO HIS FACE WITH HIS MISGOVERNMENT ORDERED THIS PSALM TO BE SUNG” in the same female voice).''}
    \label{tab:mos_text}
\end{table*}

\end{document}